\newif\ifarxiv
\arxivtrue
\documentclass[twoside,11pt]{article}
\ifarxiv
\usepackage{jair, theapa, rawfonts}
\else
\usepackage{theapa, rawfonts}
\fi

\ifarxiv
\else
\jairheading{1}{2018}{1-15}{6/18}{7/18}

\ShortHeadings{TensorLog: Deep Learning Meets Probabilistic DBs}
{Cohen and Fan}
\firstpageno{1}
\fi

\usepackage{graphicx}
\newcommand{\cd}[1]{{\small \texttt{#1}}}
\newtheorem{thm}{Theorem}

\newif\iflong
\longfalse

\newcommand{\trm}[1]{\textit{#1}}
\newcommand{\vek}[1]{\textbf{#1}}

\newcommand{\M}{\textbf{M}}
\newcommand{\exs}{{\cal E}x}
\newcommand{\model}{\textit{Model}}
\newcommand{\C}{{\cal{C}}}
\newcommand{\T}{{\cal{T}}}
\newcommand{\DB}{{\cal{DB}}}
\newcommand{\onenorm}[1]{{|\!|{#1}|\!|_1}}

\newcommand{\indicate}[1]{|\![{#1}]\!|}
\newenvironment{alg}{\begin{minipage}[t]{\textwidth}\begin{tabbing}12\=12\=12\=12\=12\=12\=12\=12\=\kill}{\end{tabbing}\end{minipage}}

\begin{document}

\title{TensorLog: Deep Learning Meets Probabilistic Databases}

\author{ \name William W. Cohen
  \name Fan Yang
  \name Kathryn Rivard Mazaitis \\
  \addr Machine Learning Department \\
  Carnegie Mellon University\\
  5000 Forbes Avenue, Pittsburgh PA 15208
}
% For research notes, remove the comment character in the line below.
% \researchnote

\maketitle

\begin{abstract}
We present an implementation of a probabilistic first-order logic
called TensorLog, in which classes of logical queries are compiled
into differentiable functions in a neural-network infrastructure such
as Tensorflow or Theano.  This leads to a close integration of
probabilistic logical reasoning with deep-learning infrastructure: in
particular, it enables high-performance deep learning frameworks to be
used for tuning the parameters of a probabilistic logic.  Experimental
results show that TensorLog scales to problems involving hundreds of
thousands of knowledge-base triples and tens of thousands of examples.
\end{abstract}

\section{Introduction}

\subsection{Motivation}

Recent progress in deep learning has profoundly affected many areas of
artificial intelligence.  One exception is probabilistic first-order
logical reasoning. In this paper, we seek to closely integrate
probabilistic logical reasoning with the powerful infrastructure that
has been developed for deep learning.  The end goal is to enable deep
learners to incorporate first-order probabilistic KBs, and conversely,
to enable probabilistic reasoning over the outputs of deep learners.

\begin{figure}
\begin{tabbing}1234\=\kill
\cd{answer(Question,Answer) :-}  \\
\> \cd{classification(Question,aboutActedIn),} \\
\> \cd{mentionsEntity(Question,Entity), actedIn(Answer,Entity).}\\
\cd{answer(Question,Answer) :- } \\
\> \cd{classification(Question,aboutDirected),} \\ 
\> \cd{mentionsEntity(Question,Entity), directed(Answer,Entity).}\\
\cd{answer(Question,Answer) :- } \\
\> \cd{classification(Question,aboutProduced), }\\
\> \cd{mentionsEntity(Question,Entity), produced(Answer,Entity).}\\
\ldots\\
\cd{mentionsEntity(Question,Entity) :- } \\
\> \cd{containsNGram(Question,NGram), matches(NGram,Name),}\\
\> \cd{possibleName(Entity,Name), \underline{popular}(Entity).}\\
~\\
\cd{classification(Question,Y) :- }\\
\> \cd{containsNGram(Question,NGram), \underline{indicatesLabel}(NGram,Y).}\\
\cd{matches(NGram,Name) :- }\\
\> \cd{containsWord(NGram,Word), containsWord(Name,Word), \underline{important}(Word).}\\
\end{tabbing}
\caption{A simple theory for question-answering against a KB.}
\label{fig:qatheory}
\end{figure}

As motivation, consider the program of Figure~\ref{fig:qatheory},
which could be plausibly used for answering simple natural-language
questions against a KB, such as ``Who was the director of Apocalyse
Now?''  The main predicate \cd{answer} takes a question and produces
an answer (which would be an entity in the KB).  The predicates
\cd{actedIn}, \cd{directed}, etc, are from the KB.  For the purpose of
performing natural-language analysis, the KB has also been extended
with facts about the text that composes the training and test data:
the KB stores information about word $n$-grams contained in the
question, the strings that are possible names of an entity, and the
words that are contained in these names and $n$-grams.  The underlined
predicates \cd{indicatesLabel}, \cd{important}, and \cd{popular} are
``soft'' KB predicates, and the goal of learning is to find
appropriate weights for the soft-predicate facts---e.g., to learn that
\cd{indicatesLabel(director, aboutDirected)} has high weight.  Ideally
these weights would be learned indirectly, from {observing inferences
  made using the KB}. In this case we would like to learn from
question-answer pairs, which rely indirectly on the KB predicates like
\cd{actedIn}, etc, rather than from hand-classified questions, or
judgements about specific facts in the soft predicates.

TensorLog, the system we describe here, makes this possible to do at
reasonable scale using conventional neural-network platforms.  For
instance, for a variant of the problem above, we can learn from 10,000
questions against a KB of 420,000 tuples in around 200 seconds per
epoch, on a typical desktop with a single GPU.

\subsection{Approach and Contributions}

The main technical obstacle to integration of probabilistic logics
into deep learners is that most existing first-order probabilistic
logics are not easily adapted to evaluation on a GPU.  One superficial
problem is that the computations made in theorem-proving are not
numeric, but there is also a more fundamental problem, which we will
now discuss.

The most common approach to first-order inference is to ``ground'' a
first-order logic by converting it to a zeroth-order format, such as a
boolean formula or a probabilistic graphical model.  For instance, in
the context of a particular KB, the rule
\begin{equation}  \label{eq:join2}
{p(X,Y) \leftarrow q(Y,Z),r(Z,Y).}
\end{equation}
can be ``grounded'' as the following finite boolean disjunction,
where $\C$ is the set of objects in the KB:
\[ \bigvee_{\exists x, y, z \in \C} 
     \left( {p}(x,y) \vee \neg {q}(y,z) \vee \neg {r}(z,y) \right) 
\]
This boolean disjunction can embedded in a neural network, e.g. to
initialize an architecture \cite{TowellAAAI90} or as a regularizer
\cite{hu2016harnessing,riedelinjecting2015}.  For probabilistic
first-order languages (e.g., Markov logic networks
\cite{RichardsonMLJ2006}), grounding typically results in a directed
graphical model (see \cite{kimmig2015lifted} for a survey of this
work).

The problem with this approach is that groundings can be very large:
even the small rule above gives a grounding of size $o(|\C|^3)$, which
is likely much larger than the size of the KB, and a grounding of size
$o(|\C|^n)$ is produced by a rule like
\begin{equation} \label{eq:chain}
{p}(X_0,X_n) \leftarrow q_1(X_0,X_1),q_2(X_1,X_2),\ldots,q_n(X_{n-1},X_n)
\end{equation}
The target architecture for modern deep learners is based on GPUs,
which have limited memory: hence the grounding approach can be used
only for small KBs and short rules. For example,
\cite{serafini2016logic} describes experimental results with five
rules and a few dozen facts, and the largest datasets considered by
\cite{DBLP:journals/corr/SourekAZK15} contain only about 3500
examples.

Although not all probabilistic logic implementations require explicit
grounding, a similar problem arises in using neural-network platforms
to implement any probabilistic logic which is computationally
hard.  For many probabilistic logics, answering queries is
\#P-complete or worse.  Since the networks constructed in modern deep
learning platforms can be evaluated in time polynomial in their size,
no polysize network can implement such a logic, unless \#P=P.

This paper addresses these obstacles with several interrelated
contributions. First, in Section~\ref{sec:background}, we identify a
restricted family of probabilistic deductive databases (PrDDBs) called
\trm{polytree-limited stochastic deductive knowledge graphs
  (ptree-SDKGs)} which are tractable, but still reasonably expressive.
This formalism is a variant of stochastic logic programs (SLPs). We
also show that ptree-SDKGs are in some sense maximally expressive, in
that we cannot drop the polytree restriction, or switch to a more
conventional possible-worlds semantics, without making inference
intractible.

Next, in Section~\ref{sec:inf-alg}, we present an algorithm for
performing inference for ptree-SDKGs. This algorithm performs
inference with a dynamic-programming method, which we formalize as
belief propagation on a certain factor graph, where each random
variable in the factor graph correspond to possible bindings to a
logical variable in a proof, and the factors correspond to database
predicates. In other words, the random variables are multinomials over
all constants in the database, and the factors constrain these
bindings to be consistent with database predicates that relate the
corresponding logical variables.  Although this is a simple idea, to
our knowledge it is novel.  We also discuss in some detail our
implementation of this logic, called TensorLog.

We finally discuss related work, experimental results, and present
conclusions.

\section{Background} \label{sec:background}

\subsection{Deductive DBs}

\begin{figure}
%\begin{small}
\begin{center}
\begin{center}
\begin{tabular}[t]{l}
1. \cd{uncle(X,Y):-child(X,W),brother(W,Y).}\\ 2. \cd{uncle(X,Y):-aunt(X,W),husband(W,Y).}\\ 3. \cd{status(X,tired):-child(W,X),infant(W).}\\
\end{tabular}\begin{tabular}[t]{ll}
\cd{child(liam,eve)} & 0.99   \\
\cd{child(dave,eve)} & 0.99   \\
\cd{child(liam,bob)} & 0.75   \\
\cd{husband(eve,bob)} & 0.9    \\
\cd{infant(liam)} & 0.7     \\
\cd{infant(dave)} & 0.1     \\
\cd{aunt(joe,eve)} & 0.9     \\
\cd{brother(eve,chip)} & 0.9 
\end{tabular}
\end{center}
\end{center}
%\end{small}
\caption{ An example database and theory.  Uppercase symbols are
  universally quantified variables, and so clause 3 should be read as
  a logical implication: for all database constants $c_X$ and $c_W$,
  if \cd{child($c_X$,$c_W$)} and \cd{infant($c_W$)} can be proved,
  then \cd{status($c_X$,tired)} can also be proved.}\label{fig:ddb}
\end{figure}

\begin{figure}
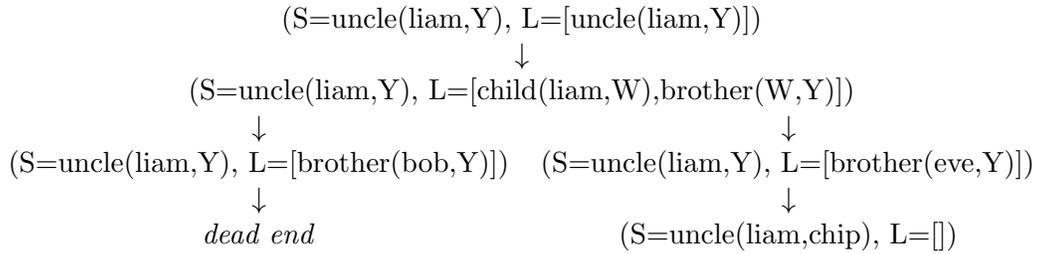

\begin{center}
\begin{tabular}{cc}
  \multicolumn{2}{c}{ (S=uncle(liam,Y), L=[uncle(liam,Y)]) }   \\
  \multicolumn{2}{c}{ $\downarrow$ } \\
  \multicolumn{2}{c}{ (S=uncle(liam,Y), L=[child(liam,W),brother(W,Y)]) } \\
  $\downarrow$ & $\downarrow$ \\
  (S=uncle(liam,Y), L=[brother(bob,Y)]) & (S=uncle(liam,Y), L=[brother(eve,Y)]) \\
  $\downarrow$ & $\downarrow$                                                   \\
\textit{dead end} & (S=uncle(liam,chip), L=[])                                  \\
\end{tabular}
\end{center}
\caption{ An example proof tree.  From root to second level uses rule 1; next level
uses unit clause \cd{child(liam,bob):-} on left and unit clause \cd{child(liam,eve):-} on right;
final level uses \cd{brother(eve,chip):-} on the right.}
\label{fig:tree}
\end{figure}

In this section we review the usual definitions for logic programs and
deductive databases, and also introduce the term \trm{deductive
  knowledge graph (DKG)} for deductive databases containing only unary
and binary predicates.  This section can be omitted by readers
familiar with logic programming.

An example of a \trm{deductive database} (DDB) is shown in
Figure~\ref{fig:ddb}.  A \trm{database}, $\DB$, is a set
$\{f_1,\ldots,f_N\}$ of ground facts. (For the moment, ignore the
numbers associated with each database fact in the figure.) A theory,
$\T$, is a set of function-free Horn clauses. Clauses are written
$A\cd{:-}B_1,\ldots,B_k$, where $A$ is called the \trm{head} of the
clause, $B_1,\ldots,B_k$ is the \trm{body}, and $A$ and the $B_i$'s
are called \trm{literals}. Literals must be of the form
$p(X_1,\ldots,X_k)$, where $p$ is a \trm{predicate symbol} and the
$X_i$'s either logical variables or database constants.  The set of
all database constants is written $\C$.  The number of arguments $k$
to a literal is called its \trm{arity}.

In this paper we focus on the case where all literals are binary or
unary, i.e., have arity no more than two.  We will call such a
database a \trm{knowledge graph (KG)}, and the program a
\trm{deductive knowledge graph (DKG)}.  We will also assume that
constants appear only in the database, not in the theory (although
this assumption can be relaxed).

Clauses can be understood as logical implications.  Let $\sigma$ be a
\trm{substitution}, i.e., a mapping from logical variables to
constants in $\C$, and let $\sigma(L)$ be the result of replacing all
logical variables $X$ in the literal $L$ with $\sigma(X)$.  A set of
tuples $S$ is \trm{deductively closed} with respect to the clause
$A\leftarrow{}B_1,\ldots,B_k$ iff for all substitutions $\sigma$,
either $\sigma(A) \in {}S$ or $\exists B_i:\sigma(B_i)\not\in{}S$.
For example, if $S$ contains the facts of Figure~\ref{fig:ddb}, $S$ is
not deductively closed with respect to the clause 1 unless it also
contains \cd{uncle(chip,liam)} and \cd{uncle(chip,dave)}.  The
\trm{least model} for a pair $\DB,\T$, written $\model(\DB,\T)$, is
the smallest superset of $\DB$ that is deductively closed with respect
to every clause in $\T$.  This least model is unique, and in the usual
DDB semantics, a ground fact $f$ is considered ``true'' iff
$f\in\model(\DB,T)$.

There are two broad classes of algorithms for inference in a DDB.
\trm{Bottom-up inference} explicitly computes the set $\model(\DB,\T)$
iteratively.  Bottom-up inference repeatedly extends a set of facts
$S$, which initially contains just the database facts, by looking for
rules which ``fire'' on $S$ and using them derive new facts.  (More
formally, one looks for rules $A\leftarrow{}B_1,\ldots,B_k$ and
substitutions $\sigma$ such that $\forall i$, $\sigma(B_i)\in S$, and
then adds the derived fact $\sigma(A)$ to $S$.)  This process is then
repeated until it converges.  For DDB programs, bottom-up inference
takes time polynomial in the size of the database $|\DB|$, but
exponential in the length of the longest clause in $\T$
\cite{ramakrishnan1995survey}.

One problem with bottom-up theorem-proving is that it explicitly
generates $\model(\DB,\T)$, which can be much larger than the original
database.  The alternative is \trm{top-down inference}.  Here, the
algorithm does not compute a least model explicitly: instead, it takes
as input a query fact $f$ and determines whether $f$ is derivable,
i.e., if $f\in\model(\DB,\T)$.  More generally, one might retrieve all
derivable facts that match some pattern, e.g., find all values of $Y$
such that \cd{uncle(joe,Y)} holds.  (Formally, given
$Q=\cd{uncle(joe,Y)}$, we would like to find all $f\in\model(\DB,\T)$
which are \trm{instances of $Q$}, where an $f$ is defined to be an
\trm{instance of $Q$} iff $\exists \sigma:f=\sigma(Q)$..

To describe top-down theorem-proving, we note that facts in the
database can also be viewed as clauses: in particular a fact $p(a,b)$
can be viewed as a clause $p(a,b)\leftarrow$ which has $p(a,b)$ as its
head and an empty body.  This sort of clause is called a \trm{unit
  clause}.  We will use $\T^{+\DB}$ to denote the theory $\T$
augmented with unit clauses for each database fact.  A top-down
theorem prover can be viewed as constructing and searching a following
tree, using the theory $\T^{+\DB}$.  The process is illustrated in
Figure~\ref{fig:tree}, and detailed below.
\begin{enumerate}
\item The root vertex is a pair $(S,L)$, where $S$ is the query $Q$, and
  $L$ is a list containing only $Q$.  In general every vertex is a pair
  where $S$ is something derived from $Q$, and $L$ is a list of
  literals left to prove.
\item For any vertex $(S,L)$, where $L=[G_1,\ldots,G_n]$, there is a
  child vertex $(S',L')$ for each rule $A\leftarrow{}B_1,\ldots,B_k \in
  \T^{+\DB}$ and substitution $\sigma$ for which
  $\sigma(G_i)=\sigma(A)$ for some $G_i$.  In this child node,
  $S'=\sigma(S)$, 
  and \[ 
L'= [\sigma(G_1),\ldots,\sigma(G_{i-1}),\sigma(B_1),\ldots,\sigma(B_k),\sigma(G_{i+1}),\ldots,\sigma(G_n)]
\]
\end{enumerate}
Note that $L'$ is smaller than $L$ if the clause selected is a unit
clause (i.e., a fact).  If $L'$ is empty, then the vertex is called a
\trm{solution vertex}. In any solution vertex $(S,L)$, if $S$ contains
no variables,\footnote{If \textit{S} does have variables in it, then
  any fact $f$ which can be constructed by replacing variables in $Q$
  with database constants is in the least model.  For clarity we will
  ignore this complication in the discussion below.} then $S$ is an
instance of $Q$ and is in $\model(\T,\DB)$.

If $\T$ is not recursive, or if recursion is limited to a fixed depth,
then the proof graph is finite.  We will restrict our discussion below
to theories with finite proof graphs.  For this case, the set of all
answers to a query $Q$ can be found by systematically searching the
proof tree for all solution vertices.  A number of strategies exist
for this, but one popular one is that used by Prolog, which uses
depth-first search, ordering edges by picking the first rule
$A\leftarrow B_1,\ldots,B_k$ in a fixed order, and only matching rules
against the first element of $L$.  This strategy can be implemented
quite efficiently and is easily extended to much more general logic
programs.

\subsection{SLPs and stochastic deductive KGs}

There are a number of approaches to incorporating probabilistic
reasoning in first-order logics.  We focus here on \trm{stochastic
  logic programs (SLPs)} \cite{DBLP:journals/ml/Cussens01}, in which
the theory $\T$ is extended by associating with each rule $r$ a
non-negative scalar weight $\theta_r$.  Below we summarize the
semantics associated with SLPs, for completeness, and refer the reader
to \cite{DBLP:journals/ml/Cussens01} for details.

In an SLP weights $\theta_r$ are added to edges of the top-down proof
graph the natural way: when a rule $r$ is used to create an edge
$(S,L)\rightarrow (S',L')'$, this edge is given weight $\theta_r$.  We
define the {weight of a path $v_1\rightarrow\ldots\rightarrow
  v_n$} in the proof graph for $Q$ to be the product of the weights of
the edges in the path, and the weight of a node $v$ to be the sum of
the weights of the paths from the root note $v_0=(Q,[Q])$ to $v$. If $r_{v,v'}$ is the rule used for the edge
from $v$ to $v'$, then the weight of $w_Q(v_n)$ is
\[
w_Q(v_n) \equiv \sum_{v_0\rightarrow\ldots\rightarrow v_n} \prod_{i=0}^{n-1} \theta_{r_{v_i,v_{i+1}}}
\]
The weight of an answer $f$ to query $Q$ is defined by summing over
paths to solution nodes that yield $f$:   
\begin{equation} \label{eq:wq}
w_Q(f) \equiv \sum_{v:v=(\textit{f},[])} w_Q(v)
\end{equation}
(Here $[]$ is the empty list, which indicates a solution vertex has
been reached.) Finally, if we assume that some answers to $Q$ do
exist, we can produce a conditional probability distribution over
answers $f$ to the query $Q$ by normalizing $w_Q$, i.e.,
\[
\Pr(f|Q) \equiv \frac{1}{Z} w_Q(f)
\]
Following the terminology of \cite{DBLP:journals/ml/Cussens01} this is
a \trm{pure unnormalized SLP}.  SLPs were originally defined
\cite{muggleton1996stochastic} for a fairly expressive class of logic
programs, namely all programs which are \trm{fail free}, in the sense
that there are no ``dead ends'' in the proof graph (i.e., from every
vertex $v$, at least one solution node is reachable).  Prior work with
SLPs also considered the special case of \trm{normalized SLPs}, in
which the weights of all outgoing edges from every vertex $v$ sum to
one.  For normalized fail-free SLPs, it is simple to modify the usual
top-down theorem prover to sample from $Pr(f|Q)$.

\subsection{Stochastic deductive KGs and discussion of SLPs}

SLPs are closely connected to several other well-known types of
probabilistic reasoners.  SLPs are defined by introducing
probabilistic choices into a top-down theorem-proving process: since
top-down theorem-proving for logic programs is analogous to program
execution in ordinary programs, SLPs can be thought of as
logic-program analogs to probabilistic programming languages like
Church \cite{goodman2012church}.  Normalized SLPs are also
conceptually quite similar to stochastic grammars, such as pCFGs,
except that stochastic choices are made during theorem-proving, rather
than rewriting a string.

Here we consider three restrictions on SLPs.  First, we restrict the
program to be in DDB form---i.e., it consists of a theory $\T$ which
contains function-free clauses, and a database $\DB$ (of unit
clauses).  Second, we restrict all predicates to be unary or binary.
Third, we restrict the clauses in the theory $\T$ to have weight 1, so
that the only meaningful weights are associated with database facts.
We call this restricted SLP a \trm{stochastic deductive knowledge
  graph (SDKG)}.

For SDKGs, a final connection with other logics can be made by
considering a logic program that has been grounded by conversion to a
boolean formulae.  One simple approach to implementing a ``soft''
extension of a boolean logic is to evaluate the truth or falsity of a
formula bottom-up, deriving a numeric confidence $c$ for each
subexpression from the confidences associated with its subparts.  For
instance, one might use the rules
\begin{eqnarray*}
c(x \wedge y) & \equiv & \min(c(x),c(y)) \\
c(x \vee y) & \equiv & \max(c(x),c(y)) \\
c(\neg x) & \equiv & 1 - c(x)
\end{eqnarray*}
This approach to implementing a soft logic is is sometimes called an
\trm{extensional} approach \cite{suciu2011probabilistic}, and it is
common in practical systems: PSL \cite{brocheler2012probabilistic}
uses an extensional approach, as do several recent neural approaches
\cite{serafini2016logic,hu2016harnessing}.

Now consider modifying a top-down prover to produce a particular
boolean formula, in which each path $v_0\rightarrow\ldots\rightarrow
v_n$ is associated with a conjunction $f_1\wedge\ldots\wedge f_m$ of
all unit-clause facts used along this path, and each answer $f$ is
associated with the disjunction of these conjunctions.  Then let us
compute the unnormalized weight $w_Q(f)$ using the rules
\begin{eqnarray*}
c(x \wedge y) & \equiv & c(x)\cdot c(y) \\
c(x \vee y) & \equiv & c(x) + c(y)
\end{eqnarray*}
(which are sufficient since no negation occurs in the formula).  This
(followed by normalization) can be shown to be equivalent to the SLP
semantics.

\subsection{Complexity of reasoning with stochastic deductive KGs}

SLPs have a relatively simple proof procedure: informally, inference
only requires computing a weighted count of all proofs for a query,
and the weight for any particular proof can be computed quickly.  A
natural question is whether computationally efficient theorem-proving
schemes exist for SLPs.  The similarity between SLPs and probabilistic
context-free grammars suggests that efficient schemes might exist,
since there are efficient dynamic-programming methods for
probabilistic parsing. Unfortunately, this is not the case: even for
the restricted case of SDKGs, computing $P(f|Q)$ is \#P-hard.

\newcommand{\statementone}{Computing $P(f|Q)$ (relative to a SDKG
  $\T,\DB$) for all possible answers $f$ of the query $Q$ is \#P-hard,
  even if there are only two such answers, the theory contains only
  two non-recursive clauses, and the KG contains only 13 facts.}

\begin{thm}\statementone
\end{thm}

A proof appears in the appendix.  The result is not especially
surprising, as it is easy to find small theories with exponentially
many proofs: e.g., the clause of Equation\ref{eq:chain} can have
exponentially many proofs, and naive proof-counting methods may be
expensive on such a clause.

Fortunately, one further restriction makes SLP theorem-proving
efficient.

For a theory clause $r=A\leftarrow B_1,\ldots,B_k$, define the
\trm{literal influence graph for $r$} to be a graph where each $B_i$
is a vertex, and there is an edge from $B_i$ to $B_j$ iff they share a
variable.  A graph is a \trm{polytree} iff there is at most one path
between any pair of vertices: i.e., if each strongly connected
component of the graph is a tree.  Finally, we define a theory to be
\trm{polytree-limited} iff the influence graph for every clause is a
polytree.  Figure~\ref{fig:factors} contains some examples of
polytree-limited clauses.

This additional restriction makes inference tractable.

\newcommand{\statementtwo}{For any SDKG with a non-recursive
  polytree-limited theory $\T$, $P(f|Q)$ can be computed in computed
  in time linear in the size of $\T$ and $\DB$.}

\begin{thm}\statementtwo
\end{thm}

The proof follows from the correctness of a dynamic-programming
algorithm for SDKG inference, which we will present below, in detail,
in Section~\ref{sec:inf-alg}. In brief, the algorithm is based on
belief propagation in a certain factor graph.  We construct a graph
where the random variables are multinomials over the set of all
database constants, and each random variable corresponds to a logical
variable in the proof graph.  The logical literals in a proof
correspond to factors, which constrain the bindings of the variables
to make the literals true.  

Importantly for the goal of compilation into deep-learning frameworks,
the message-passing steps used for belief propagation can be defined
as numerical operations, and given a predicate and an input/output
mode, the message-passing steps required to perform belief propagation
(and hence inference) can be ``unrolled'' into a function, which is
differentiable.

\subsection{Complexity of stochastic DKGs variants}

\subsubsection{Extensions that maintain efficiency}

\textit{Constants in the theory.} We will assume that constants appear
only in the database, not in the theory.  To relax this, note that it
is possible to introduce a constant into a theory by creating a
special unary predicate which holds only for that constant: e.g., to
use the constant \cd{tired}, one could create a database predicate
\cd{assign\_tired(T)} which contains the one fact
\cd{assign\_tired(tired)}, and use it to introduce a variable which is
bound to the constant \cd{tired} when needed.  For instance, the
clause 3 of Figure~\ref{fig:ddb} would be rewritten as
\begin{equation} \label{eq:tired}
\cd{status(X,T):-assign\_tired(T),child(X,W),infant(W).}
\end{equation}
Without loss of generality, we assume henceforth that constants only
appear in literals of this sort.

\textit{Rule weights and rule features.} In a SDKG, weights are
associated only with \emph{facts} in the databases, not with
\emph{rules} in the theory (which differs from the usual SLP
definition).  However, there is a standard ``trick'' which can be used
to lift weights from a database into rules: one simply introduces a
special clause-specific fact, and add it to the clause body
\cite{poole1997independent}.  For example, a weighted version of
clause 3 could be re-written as
\[
\cd{status(X,tired):-assign\_c3(RuleId),weighted(RuleId),child(W,X),infant(W)}
\]
where the (parameterized) fact \cd{weighted(c3)} appears in $\DB$, and
the constant \cd{c3} appears nowhere else in $\DB$.

In some probabilistic logics, e.g., ProPPR \cite{wang2013programming}
one can attach a computed set of features to a rule in order to weight
it: e.g., one can write 
\[ \cd{status(X,tired):-$\{$weighted(A):child(W,X),age(W,A)$\}$}
\]
which indicates that the all the ages of the children of $X$ should be
used as features to determine if the rule succeeds.  This is
equivalent to the rule \cd{status(X,tired)} :-\cd{child(W,X),
  age(W,A), weighted(A)}, and in the experiments below, where we
compare to ProPPR, we use this construction.

\subsubsection{Extension to possible-worlds semantics}

In the SLP semantics, the parameters $\Theta$ only have meaning in the
context of the set of proofs derivable using the theory $\T$.  This
can be thought of as a ``possible proofs'' semantics.  It has been
argued that it is more natural to adopt a ``possible worlds''
semantics, in which $\Theta$ is used to define a distribution,
$\Pr(I|\DB,\Theta)$, over ``hard'' databases, and the probability of a
derived fact $f$ is defined as follows, where $\indicate{\cdot}$ is a
zero-one indicator function:
\begin{equation} \label{eq:ddb-semantics}
 \Pr_\cd{TupInd}(f|\T,\DB,\Theta) \equiv \sum_{I} \indicate{f\in\model(I,\T)} \cdot \Pr(I|\DB,\Theta)
\end{equation} 
Potential hard databases are often called \trm{interpretations} in
this setting. The simplest such ``possible worlds'' model is the
\trm{tuple independence} model for PrDDB's
\cite{suciu2011probabilistic}: in this model, to generate an
interpretation $I$, each fact $f\in\DB$ sampled by independent coin
tosses, i.e., \( \Pr_\cd{TupInd}(I|\DB,\Theta) \equiv \prod_{t \in I}
\theta_t \cdot \prod_{t \in \DB-I} (1-\theta_t) \).

ProbLog \cite{fierens2016} is one well-known logic programming
language which adopts this semantics, and there is a large literature
(for surveys, see \cite{suciu2011probabilistic,de2008probabilistic})
on approaches to more tractibly estimating Eq~\ref{eq:ddb-semantics},
which naively requires marginalizing over all $2^{|\DB|}$
interpretations.  A natural question to ask is whether
polytree-limited SDKGs, which are tractible under the possible-proofs
semantics of SLPs, are also tractible under a possible-worlds
semantics.  Unfortunately, this is not the case.

\newcommand{\statementthree}{Computing $P(f)$ in the tuple-independent
  possible-worlds semantics for a single ground fact $f$ is \#P-hard.}

\begin{thm}\statementthree
\end{thm}

This result is well known: for instance, Suciu and Olteanu
\cite{suciu2011probabilistic} show that it is \#P-hard to compute
probabilities against the one-rule theory \cd{p(X,Y) :-
  q(X,Z),r(Z,Y).}  For completeness, the appendix to this paper
contains a proof, which emphasizes the fact that reasonable syntactic
restrictions (such as polytree-limited theories) are unlikely to make
inference tractible.  In particular, the theory used in the
construction is extremely simple: all predicates are unary, and
contain only three literals in their body.

\section{Efficient differentiable inference for polytree-limited SDKGs} \label{sec:inf-alg}

In this section we present an efficient dynamic-programming method for
inference in polytree-limited SDKGs. We formalize this method as
belief propagation on a certain factor graph, where the random
variables in the factor graph correspond to possible bindings to a
logical variable in a proof, and the factors correspond to database
predicates. In other words, the random variables are multinomials over
all constants in the database, and the factors will constrain these
bindings to be consistent with database predicates that related the
corresponding logical variables.  

Although using belief propagation in this way is a simple idea, to our
knowledge it is a novel method for first-order probabilistic
inference.  Certainly it is quite different from more common
formulations of first-order probabilistic inference, where random
variables typically are Bernoulli random variables, which correspond
to \emph{potential} ground database facts (i.e., elements of the
Herbrand base of the program.)

\subsection{Numeric encoding of PrDDB's and queries}

Because our ultimate goal is integration with neural networks, we will
implement reasoning by defining a series of numeric functions, each of
which finds answers to a particular family of queries.  It will be
convenient to encode the database numerically. We will assume all
constants have been mapped to integers.  For a constant $c\in\C$, we
define $\vek{u}_c$ to be a one-hot row-vector representation for $c$,
i.e., a row vector of dimension $|\C|$ where $\vek{u}[c]=1$ and
$\vek{u}[c']=0$ for $c'\not=C$.  We can also represent a binary
predicate $p$ by a sparse matrix $\M_p$, where
$\M_p[a,b]=\theta_{p(a,b)}$ if $p(a,b)\in\DB$, and a unary predicate
$q$ as an analogous row vector $\vek{v}_q$.  Note that $\M_p$ encodes
information not only about the database facts in predicate $p$, but
also about their parameter values.  Collectively, the matrices
$M_{p_1}$, \ldots, $M_{p_n}$ for the predicates $p_1,\ldots,p_n$ can
be viewed as a three-dimensional tensor.

Our main interest here is queries that retrieve all derivable facts
that match some query $Q$: e.g., to find all values of $Y$ such that
\cd{uncle(joe,Y)} holds.  We define an \trm{argument-retrieval query}
$Q$ as query of the form $p(c,Y)$ or $p(Y,c)$. We say that $p(c,Y)$
has an \trm{input-output mode} of \cd{in,out} and $p(Y,c)$ has an
input-output mode of \cd{out,in}. For the sake of brevity, below we
will assume below the mode \cd{in,out} when possible, and abbreviate
the two modes as \cd{io} and \cd{io}.

The \trm{response} to a query $p(c,Y)$ is a distribution over possible
substitutions for $Y$, encoded as a vector $\vek{v}_{Y}$ such that for
all constants $d\in\C$, $\vek{v}_{Y}[d] =
\Pr(p(c,d)|Q=p(x,Y),\T,\DB,\Theta)$.  Note that in the SLP model
$\vek{v}_{Y}$ is a conditional probability vector, conditioned of
$Q=p(c,Y)$, which we will sometimes emphasize with denoting it as
$\vek{v}_{Y|c}$.  Formally if $U_{p(c,Y)}$ the set of facts $f$ that
``match'' (are instances of) $p(c,Y)$, then
\[ \vek{v}_{Y|c}[d] =
\Pr(f=p(c,d)|f\in{}U_{p(c,Y)},\T,\DB,\Theta) \equiv \frac{1}{Z} w_Q(f=p(c,d))
\]

Although here we only consider single-literal queries, we note that
more complex queries can be answered by extending the theory: e.g., to
find 
\[ \{ \cd{Y: uncle(joe,X),husband(X,Y)}\}
\] 
we could add the clause \cd{q1(Y):-uncle(joe,X),husband(X,Y)} to the
theory and find the answer to \cd{q1(Y)}.

Since the goal of our reasoning system is to correctly answer queries
using functions, we also introduce a notation for functions that
answer particular types of queries: in particular, for a predicate
symbol $p$, $f^p_\cd{io}$ denotes a \trm{query response function} for
all queries with predicate $p$ and mode \cd{io}. We define a 
\trm{query response function} for a query of the
form $p(c,Y)$ to be a function which, when given a one-hot encoding of $c$, $f^p_\cd{io}$
returns the appropriate conditional probability vector:
\begin{equation} \label{eq:def-f}
 f^p_\cd{io}(\vek{u}_c) \equiv \vek{v}_{Y|c} 
%    \mbox{~where~} \forall d\in C: \vek{v}_{Y|c}[d] =
%             \Pr(f=p(c,d)|f\in{}U_{p(c,Y)},\T,\DB,\Theta)
\end{equation}
We analogously define $f^p_\cd{oi}$, Finally, we define $g^p_\cd{io}$
to be the unnormalized version of this function, i.e., the weight of
$f$ according to $w_Q(f)$:
\[
 g^p_\cd{io}(\vek{u}_c) \equiv w_Q(f)
\]

\begin{figure}
\centerline{\includegraphics[width=0.8\linewidth]{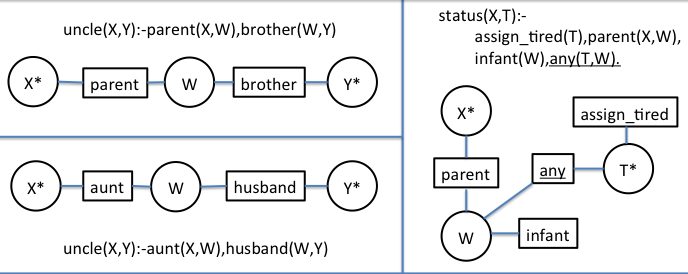}}
\caption{ Examples of factor graphs for the example
  theory.} \label{fig:factors}
\end{figure}

For convenience, we will introduce another special DB predicate
\cd{any}, where \cd{any($a,b$)} is conceptually true for any pair of
constants $a,b$; however, as we show below, the matrix $\M_\cd{any}$
need not be explicitly stored.  We also constrain clause heads to
contain distinct variables which all appear also in the body.

\subsection{Efficient inference for one-clause theories} 

We will start by considering a highly restricted class of theories
$\T$, namely programs containing only one non-recursive
polytree-limited clause $r$ that obeys the restrictions above.  We
build a factor graph $G_r$ for $r$ as follows: for each logical
variable $W$ in the body, there is a random variable $W$; and for
every literal $q(W_i,W_j)$ in the body of the clause, there is a
factor with potentials $\M_q$ linking variables $W_i$ and $W_j$.
Finally, if the factor graph is disconnected, we add \cd{any} factors
between the components until it is connected.
Figure~\ref{fig:factors} gives examples.  The variables appearing in
the clause's head are starred.

The correctness of this procedure follow immediately from the
convergence of belief propagation on factor graphs for polytrees
\cite{kschischang2001factor}.

\begin{figure}
\begin{center}
%\begin{small}
\begin{alg}
\textbf{define} compileMessage($L \rightarrow X$):\\
\> assume wolg that $L=q(X)$ or $L=p(X_i,X_o)$\\
\> generate a new variable name  $\vek{v}_{L,X}$  \\
\> \textbf{if} $L=q(X)$ \textbf{then}\\
\> \>  emitOperation( $\vek{v}_{L,X} = \vek{v}_q$)\\
\> \textbf{else if} $X$ is the output variable $X_o$ of $L$ \textbf{then}\\
\> \> $\vek{v}_i =$ compileMessage($X_i \rightarrow L$)\\
\> \> emitOperation( $\vek{v}_{L,X} = \vek{v}_i \cdot \M_p$ )\\
\> \textbf{else if} $X$ is the input variable $X_i$ of $L$ \textbf{then}\\
\> \> $\vek{v}_o =$ compileMessage($X_i \rightarrow L$) \\% 
\> \> emitOperation( $\vek{v}_{L,X} = \vek{v}_o \cdot \M_p^T$ ) \\ 
\> \textbf{return} $\vek{v}_{L,X}$\\
\end{alg}~~~~\begin{alg}
\textbf{define} compileMessage($X \rightarrow L$):  \\
\> \textbf{if} $X$ is the input variable $X$ \textbf{then}\\
\> \> \textbf{return} $\vek{u}_c$, the input\\
\> \textbf{else}\\
\> \> generate a new variable name $\vek{v}_X$\\
\> \> assume $L_1,L_2,\ldots,L_k$ are the \\
\> \> ~~ neighbors of $X$ excluding $L$ \\
\> \> \textbf{for} $i=1,\ldots,k$ \textbf{do}\\
\> \> \> $\vek{v}_i =$ compileMessage($L_i \rightarrow X$)\\
\> \> emitOperation($\vek{v}_X = \vek{v}_1 \circ \cdots \circ \vek{v}_k$) \\
\> \> \textbf{return} $\vek{v}_X$\\
\end{alg}
%\end{small}
\end{center}
\caption{ Algorithm for unrolling belief propagation on a
  polytree into a sequence of message-computation operations. Notes:
  (1) if $L=p(X_o,X_i)$ then replace $\M_p$ with $\M_p^T$ (the
  transpose). (2) Here $\vek{v}_1 \circ \vek{v}_2$ denotes the
  Hadamard (component-wise) product, and if $k=0$ an all-ones vector
  is returned.} \label{fig:alg}
\end{figure}

BP over $G_r$ can now be used to compute the conditional vectors
$f^p_\cd{io}(\vek{u}_c)$ and $f^p_\cd{oi}(\vek{u}_c)$.  For example to
compute $f^p_\cd{io}(\vek{u}_c)$ for clause 1, we would set the
message for the evidence variable $X$ to $\vek{u}_c$, run BP, and read
out as the value of $f$ the marginal distribution for $Y$.  

\subsection{Differentiable inference for one-clause theories} 

To make the final step toward integration of this algorithm with
neural-network platforms, we must finally compute an explicit,
differentiable, query response function, which computes
$f^p_\cd{io}(\vek{u}_c)$.  To do this we ``unroll'' the
message-passing steps into a series of operations.
Figure~\ref{fig:alg} shows the algorithm used in the current
implementation of TensorLog, which follows previous work in
translating belief propagation to differentiable form
\cite{gormley_approximation-aware_2015}.

In the code, we found it convenient to extend the notion of
input-output modes for a query, as follows: a variable $X$ appearing
in a literal $L=p(X,Y)$ in a clause body is an \trm{nominal input} if
it appears in the input position of the head, or any literal to the
left of $L$ in the body, and is an \trm{nomimal output} otherwise.  In
Prolog a convention is that nominal inputs appear as the first
argument of a predicate, and in TensorLog, if the user respects this
convention, then ``forward'' message-passing steps use $M_p$ rather
than $M_p^T$ (reducing the cost of transposing large $\DB$-derived
matrices, since our message-passing schedule tries to maximize forward
messages.)  The code contains two mutually recursive routines, and is
invoked by requesting a message from the output variable to a
fictional output literal. The result will be to emit a series of
operations, and return the name of a register that contains the
unnormalized conditional probability vector for the output variable.
For instance, for the sample clauses, the functions returned are shown
in Table~\ref{tab:messages}.

\begin{table}
\begin{tabular}{c|l|l|l} 
\hline
  Rule        & r1: uncle(X,Y):-                    & r2: uncle(X,Y):-                              & r3: status(X,T):- \\
              & ~parent(X,W),                       & ~aunt(X,W),                                   & ~assign\_tired(T), \\
              & ~brother(W,Y)                       & husband(W,Y)                                  & ~parent(X,W),\\
              &                                     &                                               & ~infant(W),any(T,W)\\
\hline
Function & $g^{r1}_\cd{io}(\vec{u}_c)$           & $g^{r2}_\cd{io}(\vec{u}_c)$                    & $g^{r3}_\cd{io}(\vec{u}_c)$ \\
\hline
           &  $\vek{v}_{1,W} = \vek{u}_c \M_\cd{parent}$ & $\vek{v}_{1,W} = \vek{u}_c \M_\cd{aunt}$     & $\vek{v}_{2,W} = \vek{u}_c \M_\cd{parent}$ \\
Operation  & $\vek{v}_W = \vek{v}_{1,W}$                & $\vek{v}_W = \vek{v}_{1,W}$                  & $\vek{v}_{3,W} = \vek{v}_\cd{infant}$ \\
sequence   & $\vek{v}_{2,Y} = \vek{v}_W \M_\cd{brother}$ & $\vek{v}_{2,Y} = \vek{v}_W \M_\cd{husband}$  & $\vek{W} = \vek{v}_{2,W} \circ \vek{v}_{3,W}$ \\
defining   & $\vek{v}_Y = \vek{v}_{2,Y}$                & $\vek{v}_Y = \vek{v}_{2,Y}$                  & $\vek{v}_{1,T} = \vek{v}_\cd{assign\_tired}$ \\
function   &                                           &                                            & $\vek{v}_{4,T} = \vek{v}_{W} \M_\cd{any}$ \\
           &                                           &                                            & $\vek{T} = \vek{v}_{1,T} \circ \vek{v}_{4,T}$  \\
\hline
Returns    & $\vek{v}_{Y}$                              & $\vek{v}_{Y}$                               & $\vek{v}_{T}$  \\
\hline
\end{tabular}
\caption{Chains of messages constructed for the three sample clauses
  shown in Figure~\ref{fig:factors}, written as functions in pseudo
  code.}
\label{tab:messages}
\end{table}

Here we use $g^r_\cd{io}(\vec{u}_c)$ for the unnormalized version of
the query response function build from $G_r$.  One could normalize as
follows:
\begin{equation} \label{eq:normalize}
 f^p_\cd{io}(\vec{u}_c) \equiv  g^r_\cd{io}(\vec{u}_c)/\onenorm{g^r_\cd{io}(\vec{u}_c)} 
\end{equation}
where $r$ is the one-clause theory defining $p$.

\subsection{Multi-clause programs} \label{sec:multi-clause}

We now extend this idea to theories with many clauses.  We first note
that if there are several clauses with the same predicate symbol in
the head, we simply sum the unnormalized query response functions:
e.g., for the predicate \cd{uncle}, defined by rules $r_1$ and $r_2$,
we would define
\[ g^\cd{uncle}_\cd{io} = g^{r1}_\cd{io} +  g^{r2}_\cd{io}
\]
This is equivalent to building a new factor graph $G$, which would be
approximately $\cup_i G_{ri}$, together global input and output
variables, plus a factor that constrains the input variables of the
$G_{ri}$'s to be equal, plus a factor that constrains the output
variable of $G$ to be the sum of the outputs of the $G_{ri}$'s.

A more complex situation is when the clauses for one predicate, $p$,
use a second theory predicate $q$, in their body: for example, this
would be the case if \cd{aunt} was also defined in the theory, rather
than the database.  For a theory with no recursion, we can replace the
message-passing operations $\vek{v}_Y = \vek{v}_X \M_q$ with the
function call $\vek{v}_Y = g^q_\cd{io}(\vek{v}_X)$, and likewise the
operation $\vek{v}_Y = \vek{v}_X \M_q^T$ with the function call
$\vek{v}_Y = g^q_\cd{oi}(\vek{v}_X)$.  It can be shown that this is
equivalent to taking the factor graph for $q$ and ``splicing'' it into
the graph for $p$.

It is also possible to allow function calls to recurse to a fixed
maximum depth: we must simply add an extra argument that tracks depth
to the recursively-invoked $g^q$ functions, and make sure that $g^p$
returns an all-zeros vector (indicating no more proofs can be found)
when the depth bound is exceeded.  Currently this is implemented by
marking learned functions $g$ with the predicate $q$, a mode, and a
depth argument $d$, and ensuring that function calls inside
$g^p_{\cd{io},d}$ to $q$ always call the next-deeper version of the
function for $q$, e.g., $g^q_{\cd{io},d+1}$.

Computationally, the algorithm we describe is quite
efficient. Assuming the matrices $\M_p$ exist, the additional memory
needed for the factor-graph $G_r$ is linear in the size of the clause
$r$, and hence the compilation to response functions is linear in the
theory size and the number of steps of BP.  For ptree-SDKGs, $G_r$ is
a tree, the number of message-passing steps is also linear.  Message
size is (by design) limited to $|\C|$, and is often smaller in
practice, due to sparsity or type restrictions (discussed below).

\subsection{Implementation: TensorLog}

\textit{Compilation and execution.}  The current implementation of
TensorLog operates by first ``unrolling'' the belief-propagation
inference to an intermediate form consisting of sequences of abstract
operators, as suggested by the examples of Table~\ref{tab:messages}.
The ``unrolling'' code performs a number of optimizations to the
sequence in-line: one important one is to use the fact that \(
\vek{v}_X \circ (\vek{v}_Y \M_\cd{any}) = \vek{v}_X
\onenorm{\vek{v}_Y} \) to avoid explicitly building $\M_\cd{any}$.
These abstract operator sequences are then ``cross-compiled'' into
expressions on one of two possible ``back end'' deep learning
frameworks, Tensorflow \cite{abadi2016tensorflow} and Theano
\cite{bergstra2010theano}.  The operator sequences can also be
evaluated and differentiated on a ``local infrastructure'' which is
implemented in the SciPy sparse-matrix package \cite{jones2014scipy},
which includes only the few operations actually needed for inference,
and a simple gradient-descent optimizer.

The local infrastructure's main advantage is that it makes more use of
sparse-matrix representations.  In all the implementations, the
matrices that correspond to KB relations are sparse.  The messages
corresponding to a one-hot variable binding, or the possible bindings
to a variable, are sparse vectors in the local infrastructure, but
dense vectors in the Tensorflow and Theano versions, to allow use of
GPU implementations of multiplication of dense vectors and sparse
matrices.  (The implementation also supports grouping examples into
minibatches, in which case the dense vectors become dense matrices
with a number of rows equal to minibatch size.)

TensorLog compiles query response functions on demand, i.e., only as
needed to answer queries or train.  In TensorLog the parameters
$\Theta$ are partitioned by the predicate they are associated with,
making it possible to learn parameters for any selected subset of
database predicates, while keeping the remainder fixed.

\textit{Typed predicates.}
One practically important extension to the language for the Tensorflow
and Theano targets was include machinery for declaring types for the
arguments of database predicates, and inferring these types for logic
programs: for instance, for the sample program of
Figure~\ref{fig:qatheory}, one might include declarations like
\cd{actedIn(actor,film)} or \cd{indicatesLabel(ngram,questionLabel)}.
Typing reduces the size of the message vectors by a large constant
factor, which increases the potential minibatch size and speeds up
run-time by a similar factor.

\begin{figure}
\begin{small}
\begin{tabbing}1234\=1234\=1234\=1234\=1234\=1234\=1234\=\kill
tlog = tensorlog.simple.Compiler(db=''data.db'', prog=''rules.tlog'') \\
train\_data = tlog.load\_dataset(''train.exam'') \\
test\_data = tlog.load\_dataset(''test.exam'') \\
\# \textit{data is stored dictionary mapping a function specification, like $p_\cd{io}$,} \\
\# \textit{to a pair X, Y. The rows of X are possible inputs $f^p_\cd{io}$, and the rows of} \\
\# \textit{Y are desired outputs.} \\
function\_spec = train\_data.keys()[0] \\ \textit{\# assume only one function spec} \\
X,Y = train\_data[function\_spec] \\
~\\
\textit{\# construct a tensorflow version of the loss function, and function used for inference} \\
unregularized\_loss = tlog.loss(function\_spec) \\
f = tlog.inference(function\_spec) \\
\# \textit{add regularization terms to the loss} \\
regularized\_loss = unregularized\_loss\\
for weight in tlog.trainable\_db\_variables(function\_spec): \\
\> regularized\_loss = regularized\_loss + tf.reduce\_sum(tf.abs(weights))*0.01 \# \textit{L1 penalty}\\
~\\
\# \textit{set up optimizer and inputs to the optimizer}\\
optimizer = tf.train.AdagradOptimizer(rate) \\
train\_step = optimizer.minimize(regularized\_loss)\\
\# \textit{inputs are a dictionary, with keys that name the appropriate variables used in the loss function}\\
train\_step\_input = \{\} \\
train\_step\_input[tlog.input\_placeholder\_name(function\_spec)] = X \\
train\_step\_input[tlog.target\_output\_placeholder\_name(function\_spec)] = Y \\
~\\
\# \textit{run the optimizer for 10 epochs}\\
session = tf.Session()\\
session.run(tf.global\_variables\_initializer())\\
for i in range(10):\\
\>session.run(train\_step, feed\_dict=train\_step\_input)\\
~\\
\# \textit{now run the learned function on some new data}\\
result = session.run(f, feed\_dict=\{tlog.input\_placeholder\_name(function\_spec): X2\})
\end{tabbing}
\end{small}
\caption{Sample code for using TensorLog within Tensorflow.  This code minimizes an alternative version
  of the loss function which includes and L1 penalty of the weights.} \label{fig:tfcode}
\end{figure}

\textit{Constraining the optimizer.}  TensorLog's learning changes the
numeric score $\theta_f$ of every soft KG fact $f$ using gradient
descent.  Under the proof-counting semantics used in TensorLog, a fact
with a score of $\theta_f>1$ could be semantically meaningful: for
instance for $f=\cd{costar(ginger\_rogers,fred\_astaire)}$ one might
plausibly set $\theta_f$ to the number of movies those actors appeared
in together.  However it is not semantically meaningful to allow
$\theta_f$ to be negative.  To prevent this, before learning, for each
KG parameter $\theta_f$, we replace each occurrence of $\theta_f$ with
$h(\tilde{\theta}_f)$ for the function $h=\ln(1+e^x)$ (the
``softplus'' function), where $\tilde{\theta}_f \equiv
h^{-1}(\theta_f)$.  Unconstrained optimization is then performed to
optimize the value of $\tilde{\theta}_f$ to some $\tilde{\theta}^*_f$
After learning, we update $\theta_f$ to be $h(\tilde{\theta}_f^*)$,
which is always non-negative.

\textit{Regularization.}  By default, TensorLog trains to minimize
unregularized cross-entropy loss.  (Following common practice deep
learning, the default loss function replaces the conventional
normalizer of Equation~\ref{eq:normalize} with a softmax
normalization.)  However, because modern deep-learning frameworks are
quite powerful, it is relatively easy to use the cross-compiled
functions produced by TensorLog in slight variants of this learning
problem---often this requires only a few lines of code.  For instance,
Figure~\ref{fig:tfcode} illustrates how to add L1-regularization to
TensorLog's loss function (and then train) using the Tensorflow
backend.

\textit{Extension to multi-objective learning.} It is also relatively
easy to extend TensorLog in other ways.  We will discuss several
possible extensions which we have not, as yet, experimented with
extensively, although we have verified that all can be implemented in
the current framework.

\begin{sloppypar}For learning, TensorLog's training data consists of a set of
queries $p(c_1,Y),\ldots,p(c_m,Y)$, and a corresponding set of desired
outputs $\vek{v}_{Y|c_1},\ldots,\vek{v}_{Y|c_m}$. It is possible to
train with examples of multiple predicates: for instance, with the
example program of Figure~\ref{fig:qatheory}, one could include
training examples for both \cd{answer} and \cd{matches}.
\end{sloppypar}

\textit{Alternative semantics for query responses.}  One natural
extension would address a limitation of the SLP semantics, namely,
that the weighting of answers relative to a query sometimes leads to a
loss of information.  For example, suppose the answers to
\cd{father(joe,Y)} are two facts \cd{father(joe,harry)} and
\cd{father(joe,fred)}, each with weight 0.5.  This answer does not
distinguish between a world in which \cd{joe}'s paternity is
uncertain, and a world in which \cd{joe} has two fathers.  One
possible solution is to learn parameters that set an appropriate soft
threshold on each element of $w_Q$, e.g., to redefine $f^p$ as
\[ f^p(\vek{u})) = \textit{sigmoid}(g^p(\vek{u}) + b^p)
\]
where $b^p$ is a bias term.  The code required to do this
for Tensorflow is below:
\begin{tabbing}1234\=\kill
\>target = tlog.target\_output\_placeholder(function\_spec)\\
\>g = tlog.proof\_count(function\_spec) \# \textit{$g^p$, computes $w_Q$}\\
\>bias = tf.Variable(0.0, dtype=tf.float32, trainable=True) \\
\>f = tf.sigmoid(g + bias) \# \textit{function used for inference} \\
\>unregularized\_loss = tf.nn.sigmoid\_cross\_entropy\_with\_logits(g+bias,target)
\end{tabbing}
This extension illustrates an advantage of being able to embed
TensorLog inferences in a deep network.

\textit{Extension to call out to the host infrastructure.}  A second
extension is to allow TensorLog functions to ``call out'' to the
backend language.  Suppose, for example, we wish to replace the
\cd{classification} predicate in the example program of
Figure~\ref{fig:qatheory} with a Tensorflow model, e.g., a multilayer
perceptron, and that \cd{buildMLP(q)} is function that constructs a an
expression which evaluates the MLP on input \cd{q}.  We can instruct
the compiler to include this model in place of the usual function
$g^\cd{classification}_\cd{io}$ as follows:

\begin{tabbing}1234\=\kill
\>plugins = tensorlog.program.Plugins() \\
\>plugins.define(''classification/io'', buildMLP) \\
\>tlog = simple.Compiler(db=''data.db'', prog=''rules.tlog'', plugins=plugins) \\
\end{tabbing}

To date we have not experimentally explored this capability in depth;
however, it would appear to be very useful to be able to write logical
rules over arbitrary neurally-defined low-level predicates, rather
than merely over KB facts.  We note that the compilation approach also
makes it easy to export a TensorLog predicate (e.g., the \cd{answer}
predicate defined by the logic) to a deep learner, as a function which
maps a question to possible answers and their confidences. This might
be useful in building a still more complex model non-logical model
(e.g., a dialog agent which makes use of question-answering as a
subroutine.)

\section{Related Work}

\subsection{Hybrid logical/neural systems} There is a long tradition
of embedding logical expressions in neural networks for the purpose of
learning, but generally this is done indirectly, by conversion of the
logic to a boolean formula, rather than developing a differentiable
theorem-proving mechanism, as considered here.  Embedding logic may
lead to a useful architecture \cite{TowellAAAI90} or regularizer
\cite{riedelinjecting2015,hu2016harnessing}.

More recently \cite{rocktaschel2016learning} have proposed a
differentiable theorem prover, in which a proof for an example is
unrolled into a network.  Their system includes
representation-learning as a component, as well as a
template-instantiation approach (similar to \cite{wang2014structure}),
allowing structure learning as well.  However, published experiments
with the system been limited to very small datasets.  Another recent
paper \cite{DBLP:journals/corr/AndreasRDK16} describes a system in
which non-logical but compositionally defined expressions are
converted to neural components for question-answering tasks.

\subsection{Explicitly grounded probabilistic first-order languages}

Many first-order probabilistic models are implemented by
``grounding'', i.e., conversion to a more traditional representation.
In the context of a deductive DB, a rule can be considered as a finite
disjunction over ground instances: for instance, the rule
\[ \cd{p(X,Y) :- q(Y,Z),r(Z,Y). }
\]
is equivalent to 
\[ \exists x \in \C, y \in \C, x \in \C : \cd{p}(x,y) \vee \neg \cd{q}(y,z) \vee \neg \cd{r}(Z,Y) 
\]

For example, Markov logic networks (MLNs)
are a widely-used probabilistic first-order model
\cite{RichardsonMLJ2006} in which a Bernoulli random variable is
associated with each \emph{potential} ground database fact (e.g., in
the binary-predicate case, there would be a random variable for each
possible $p(a,b)$ where $a$ and $b$ are any facts in the database and
$p$ is any binary predicate) and each ground instance of a clause is a
factor.  The Markov field built by an MLN is hence of size $O(|\C|^2)$
for binary predicates, which is much larger than the factor graphs
used by TensorLog, which are of size linear in the size of the theory.
In our experiments we compare to ProPPR, which has been elsewhere
compared extensively to MLNs.

Inference on the Markov field can also be expensive, which motivated
the development of probabilistic similarity logic (PSL),
\cite{brocheler2012probabilistic} a MLN variant which uses a more
tractible hinge loss, as well as lifted relational neural networks
\cite{DBLP:journals/corr/SourekAZK15} and logic tensor networks
\cite{serafini2016logic} two recent models which grounds first-order
theories to a neural network.  However, any grounded model for a
first-order theory can be very large, limiting the scalability of such
techniques.

\subsection{Stochastic logic programs and ProPPR} \label{sec:SLPs}
As noted above, TensorLog is very closely related to stochastic logic
programs (SLPs) \cite{DBLP:journals/ml/Cussens01}. In an SLP, a
probabilistic process is associated with a top-down theorem-prover:
i.e., each clause $r$ used in a derivation has an assocated
probability $\theta_{r}$.  Let $N(r,E)$ be the number of times $r$ was
used in deriving the explanation $E$: then in SLPs, \( \Pr_\cd{SLP}(f)
= \frac{1}{Z} \sum_{E\in\exs(f)} \prod_r \theta_r^{N(r,E)} \).  The
same probability distribution can be generated by TensorLog if (1) for
each rule $r$, the body of $r$ is prefixed with the literals
\cd{assign(RuleId,$r$),weighted(RuleId)}, where $r$ is a unique
identifier for the rule and (2) $\Theta$ is constructed so that
$\theta_f=1$ for ordinary database facts $f$, and
$\theta_\cd{weighted(r)}=\theta'_\cd{r}$, where $\Theta'$ is the
parameters for a SLP.

SLPs can be \trm{normalized} or \trm{unnormalized}; in normalized
SLPs, $\Theta$ is defined so for each set of clauses $S_p$ of clauses
with the same predicate symbol $p$ in the head, $\sum_{r\in{}S_p}
\theta_r=1$.  TensorLog can represent both normalized and unnormalized
SLPs (although clearly learning must be appropriately constrained to
learn parameters for normalized SLPs.)  Normalized SLPs generalize
probabilistic context-free grammars, and unnormalized SLPs can express
Bayesian networks or Markov random fields
\cite{DBLP:journals/ml/Cussens01}.

ProPPR \cite{wang2013programming} is a variant of SLPs in which (1)
the stochastic proof-generation process is augmented with a reset, and
(2) the transitional probabilities are based on a normalized
soft-thresholded linear weighting of features.  The first extension to
SLPs can be easily modeled in TensorLog, but the second cannot: the
equivalent of ProPPR's clause-specific features can be incorporated,
but they are globally normalized, not locally normalized as in ProPPR.
%Hence ProPPR and TensorLog have a relationship somewhat analogous to
%MEMMs \cite{McCallumML2000} and CRFs \cite{LaffertyML2001}.

ProPPR also includes an approximate grounding procedure which
generates networks of bounded size.  Asymptotic analysis suggests that
ProPPR should be faster for very large database and small numbers of
training examples (assuming moderate values of $\epsilon$ and $\alpha$
are feasible to use), but that TensorLog should be faster with large
numbers of training examples and moderate-sized databases.

\section{Experiments}

\newcommand{\bst}[1]{\textbf{#1}}

\begin{table}
\begin{center}
\begin{tabular}{l|rr}
\hline
          & \multicolumn{2}{c}{Social Influence Task} \\
\hline
ProbLog2  &   20 nodes & 40-50 sec \\
TensorLog & 3327 nodes & \bst{9.2 msec} \\
\hline
\end{tabular}

\caption{Comparison to ProbLog2 on the ``friends and smokers''
  inference task.} \label{tab:smokers}

~\\
~\\

\begin{tabular}[c]{l|rrr}
\hline
          & \multicolumn{3}{c}{Path-finding} \\
          & \multicolumn{1}{c}{Size} & \multicolumn{1}{c}{Time} & \multicolumn{1}{c}{Acc} \\
\hline
ProbLog2  & 16x16 grid, $d=10$ & 100-120 sec & \\
\hline
TensorLog & 16x16 grid, $d=10$ & \bst{2.1 msec}  &    \\
          & 64x64 grid, $d=99$ & {2.2 msec}  &    \\
\cline{2-3}
\textit{trained} 
          & 16x16 grid, $d=10$ & 6.2 msec  &   99.89\% \\
%         &  \multicolumn{3}{c}{\textit{After training}} \\
\hline
\end{tabular}\begin{minipage}[c]{0.3\linewidth}
\includegraphics[width=\linewidth]{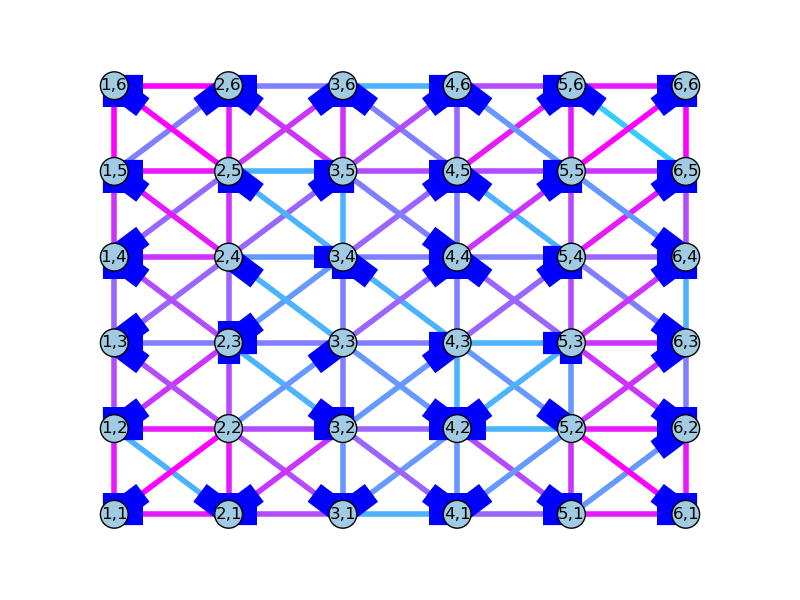}
\end{minipage}

\caption{Comparison to ProbLog2 on path-finding in a
  grid.} \label{tab:grid}

~\\
~\\
\begin{tabular}{cc|cc|cc|cc}
\hline
Grid Size & Max Depth & \multicolumn{2}{c}\# Graph Nodes & \multicolumn{2}{c}{Acc} & \multicolumn{2}{c}{Time (30 epochs)} \\
          &           & Local  & TF    & Local & TF             & Local  & TF     \\
\hline
16 & 10 & 68  & 2696 & \bst{99.9} &  97.2  &   37.6 sec &    \bst{1.1 sec} \\
18 & 12 & 80  & 3164 &  93.9 &  \bst{96.9}  &  126.1 sec &    \bst{1.8 sec}\\
20 & 14 & 92  & 3632 &  25.2 &  \bst{99.1}  &  144.9 sec &    \bst{2.8 sec}\\
22 & 16 & 104 & 4100 &   8.6 &  \bst{98.4}  &   83.8 sec &    \bst{4.2 sec}\\
24 & 18 & 116 & 4568 &   \bst{2.4} &  0.0   &  611.7 sec &    \bst{6.3 sec}\\
\hline
\end{tabular}

\caption{Learning for the path-finding task with local and Tensorflow
  (TF) backends.} \label{tab:gridlearn}

~\\
~\\

\begin{tabular}{l|r|rc} \hline
                      & \multicolumn{1}{c|}{ProPPR} & \multicolumn{1}{c}{TensorLog}  \\
\hline
CORA (13k facts,10 rules)
& AUC 83.2               & AUC \bst{97.6}   \\
\hline
UMLS (5k facts, 226 rules) 
& acc 49.8                 & acc \bst{52.5}   \\
\hline
Wordnet (276k facts)  &          &         \\
~~Hypernym (46 rules) & acc \bst{93.4}     & acc 93.3   \\
~~Hyponym (46 rules)  & acc 92.1     & acc \bst{92.8}    \\
\hline
\end{tabular}

\caption{Comparison to ProPPR on relational learning
  tasks.} \label{tab:proppr}
\end{center}
\end{table}

\subsection{Inference tasks} \label{sec:timing}

We compared TensorLog's inference time (using the local
infrastructure) with ProbLog2, a mature probabilistic logic
programming system which implements the tuple independence semantics,
on two inference problems described in \cite{fierens2016}.  One is a
version of the ``friends and smokers'' problem, a simplified model of social
influence.  In \cite{fierens2016} small graphs were artificially
generated using a preferential attachment model, the details of which
were not described; instead we used a small existing network
dataset\footnote{The Citeseer dataset from
  \cite{DBLP:conf/asunam/LinC10}.} which displays
preferential-attachment statistics.  The inference times we report are
for the same inference tasks, for a subset of 120 randomly-selected
entities.  As shown in Table~\ref{tab:smokers}, in spite of querying
six times as many entities, TensorLog is many times faster.

We also compare on a path-finding task from \cite{fierens2016}, which
is intended to test performance on deeply recursive tasks.  The goal
here is to compute fixed-depth transitive closure on a grid: in
\cite{fierens2016} a 16-by-16 grid was used, with a maximum path
length of 10.  Again TensorLog shows much faster performance, and
better scalability, as shown in Table~\ref{tab:grid} by run times on a
larger 64-by-64 grid. We set TensorLog's maximum path length to 99 for
the larger grid.

\subsection{Learning Tasks}

We also compared experimentally with ProPPR on several standard
benchmark learning tasks.  We chose two traditional relational
learning tasks on which ProPPR outperformed plausible competitors,
such as MLNs.  One was the CORA citation-matching task (from
\cite{wang2013programming}) with hand-constructed rules.\footnote{We
  replicated the experiments with the most recent version of ProPPR,
  obtaining a result slightly higher than the 2013 version's published
  AUC of 80.0}.  A second was learning the most common relation,
``affects'', from UMLS, using a rule set learned by the algorithm of
\cite{wang2014structure}.  Finally, motivated by recent comparisons
between ProPPR and embedding-based approaches to knowledge-base
completion \cite{Wang-Cohen:2016:IJCAI}, we also compared to ProPPR on
two relation-prediction tasks involving WordNet, again using rules
from the (non-recursive) theories used in
\cite{Wang-Cohen:2016:IJCAI}.

In all of these tasks parameters are learned on a separate training
set.  For TensorLog's learner, we used the local infrastructure with
the default loss function (unregularized cross-entropy loss), using a
fixed-rate gradient descent learner with the learning rate to 0.1, and
30 epochs.\footnote{Thirty epochs approximately matches ProPPR's
  runtime on a single-threaded machine.}  We also used the default
parameters for ProPPR's learning.

Table~\ref{tab:proppr} shows that the accuracy of the two systems
after learning is quite comparable, even with a rather simplistic
learning scheme.  ProPPR, of course, is not well suited to tight
integration with deep learners.

\subsection{Path-finding after learning}  

The results of Section~\ref{sec:timing} demonstrate that TensorLog's
approximation to ProbLog2's semantics is efficient, but not that it is
useful.  To demonstrate that TensorLog can efficiently and usefully
approximate deeply recursive concepts, we posed a learning task on the
16-by-16 grid, with a maximum depth of 10, and trained TensorLog to
approximate the distribution for this task.  The dataset consists of
256 grid cells connected by 2116 edges, so there are 256 example
queries of the form \cd{path(a,X)} where $a$ is a particular grid
cell.  We picked 1/3 of these queries as test, and the remainder as
train, and trained so that that the single positive answer to the
query \cd{path(a,X)} is the extreme corner closest to \cd{a}---i.e.,
one of the corners (1,1), (1,16), (16,1) or (16,16).  We set the
initial weights of the edges uniformly to 0.2.

Training for 30 epochs with the local backend and a fixed-rate
gradient descent learner, using a learning rate of 0.01, brings the
accuracy from 0\% to 99.89\% for test cases (averaged over 10 trials,
with different train/test splits). Learning takes less than 1.5
sec/epoch.  After learning query times are still quite fast, as shown
in the table.

The table also includes a visualization of the learned weights for a
small 6x6 grid.  For every pair of adjacent grid cells $u,v$, there
are two weights to learn, one for the edge from $u$ to $v$ and one for
its converse.  For each weight pair, we show a single directed edge
(the heavy blue squares are the arrows) colored by the magnitude of
the difference.

We observe that ProbLog2, in addition to implementing the full
tuple-independence semantics, implements a much more expressive logic
than considered here, including a large portion of full Prolog, while
in contrast TensorLog includes only a subset of Datalog.  So to some
extent this comparison is unfair.

We also observe that although this task seems simple, it is quite
difficult for probabilistic logics, because of deeply recursive
theories lead to large, deep proofs.  While TensorLog's inference
schemes scale well on this task, is still challenging to optimize the
parameters, especially for larger grid sizes.  One problem is that
unrolling the inference leads to very large graphs, especially after
they are compiled to the relatively fine-grained operations used in
deep-learning infrastructure.  Table~\ref{tab:gridlearn} shows the
size of the networks after compilation to Tensorflow for various
extensions of the 16-by-16 depth 10 task.  Although growth is linear
in depth, the constants are large: e.g., the Tensorflow 64-by-64 depth
99 network does not fit in memory for a 4Gb GPU.

A second problem is that the constructed networks are very deep, which
leads to problems in optimization.  For the smaller task, the local
optimizer (which is a fixed-rate gradient descent method) required
careful tuning of the initial weights and learning rate to reliably
converge.  

The size and complexity of this task suggested a second set of
experiments, where we varied the task complexity, while fixing the
parameters of two optimizers. For the local optimizer we fixed the
parameters to those used the 16-by-16 depth 10 task, and for the
Tensorflow backend, we used a the \cd{Adagrad\-Optimizer} with a
default learning rate of 1.0, running for 30 epochs. The results are
shown in Table~\ref{tab:gridlearn} (averaged over 10 trials for each
datapoint), and they illustrate several of the advantages of using a
mature deep-learning framework as the backend of TensorLog.
\begin{itemize}
\item In general learning is many times faster for the Tensorflow
  backend, which uses a GPU processor, than using the local
  infrastructure.\footnote{Learning times for the local infrastructure
    are quite variable for the larger sizes, because numerical
    instabilities often cause the optimizer to fail.  In computing
    times we discard runs where there is overflow but not when there
    is underflow, which is harder to detect.  The high variance
    accounts for the anomolously low average time for grid size 22.}

\item Although they do not completely eliminate the need for
  hyperparameter tuning, the more sophisticated optimizers available
  in Tensorflow do appear to be more robust. In particular, Adagrad
  performs well up to a depth of around 16, while the fixed-rate optimizer
  performs well only for depths 10 and 12.
\end{itemize}
We conjecture that good performance on larger grid sizes would require
use of gradient clipping.

\begin{table}
\begin{center}
\begin{tabular}{cc|cc|ccc}
\hline
\multicolumn{2}{c|}{Original KB} & \multicolumn{2}{c|}{Extended KB} & \multicolumn{3}{c}{Num Examples} \\
Num Tuples & Num Relations    &  Num Tuples & Num Relations       &  Train & Devel & Test \\
\hline
    421,243 & 10              & 1,362,670 & 12              & 96,182 & 20,000 & 10,000 \\
\hline
\end{tabular}

\caption{Statistics concerning the WikiMovies dataset.}

~\\
~\\

\begin{tabular}{lcr}
\hline
Method & Accuracy & Time per epoch \\
\hline
Subgraph/question embedding & 93.5\% & \\
Key-value memory network    & 93.9\% &  \\
\hline
TensorLog (1,000 training examples)                                  & 89.4\% &     6.1 sec \\
TensorLog (10,000 training examples)                                 & 94.8\% &     1.7 min \\
TensorLog (96,182 training examples)                                 & \bst{95.0\%} &    49.5 min \\
\hline
\end{tabular}

\end{center}
\caption{Experiments with the WikiMovies dataset.  The first two
  results are taken from
  \cite{miller-EtAl:2016:EMNLP2016}.} \label{tab:wikimovies}
\end{table} 

\subsection{Answering Natural-Language Questions Against a KB}

As larger scale experiment, we used the WikiMovies question-answering
task proposed by \cite{miller-EtAl:2016:EMNLP2016}.  This task is
similar to the one shown in Figure~\ref{fig:qatheory}.  The KB
consists of over 420k tuples containing information about 10 relations
and 16k movies.  Some sample questions with their answers are below,
with double quotes identifying KB entities.
\begin{itemize}
\item Question: Who acted in the movie Wise Guys? \\
\textit{Answers: ``Harvey Keitel'', ``Danny DeVito'', ``Joe Piscopo'', \ldots}
\item Question: what is a film written by Luke Ricci? \\
\textit{Answer: ``How to be a Serial Killer''}
\end{itemize}
We encoded the questions into the KB by extending it with two
additional relations: \cd{mentionsEntity(Q,E)}, which is true if
question \cd{Q} mentions entity \cd{E}, and \cd{hasFeature(Q,W)},
which is true if question \cd{Q} contains feature \cd{W}. The entities
mentioned in a question were extracted by looking for every longest
match to a name in the KB.  The features of a question are simply the
words in the question (minus a short stoplist).

The theory is a variant of the one given as an example in
Figure~\ref{fig:qatheory}.  The main difference is that because the
simple longest-exact-match heuristic described above identifies
entities accurately for this dataset, we made \cd{mentionsEntity} a
hard KB predicate.  We also extended the theory to handle questions
with answers that are either movie-related entities (like the actors
in the first example question) or movies (as in the second example.
Finally, we simplified the question-classification step slightly.  The
final theory contains two rules and two ``soft'' unary relations
\cd{QuestionType$_{R,1}$}, \cd{indicates\-Question\-Type$_{R,2}$} for
each relation $R$ in the original movie KB.  For example, for the
relation \cd{directedBy} the theory has the two rules

\begin{tabbing}1234\=1234\=\kill
\>\cd{answer(Question,Movie) :-} \\
\> \> \cd{mentionsEntity(Question,Entity), directedBy(Movie,Entity),}\\
\> \> \cd{hasFeature(Question,Word), indicatesQuestionType$_{\rm directedBy,1}$(Word)}\\
\>\cd{answer(Question,Entity) :-} \\
\> \> \cd{mentionsEntity(Question,Movie), directedBy(Movie,Entity),}\\
\> \> \cd{hasFeature(Question,Word), indicatesQuestionType$_{\rm directedBy,2}$(Word)}\\
\end{tabbing}

The last line of each rule acts as a linear classifier for that rule.

For efficiency we used three distinct types of entities (question ids,
entities from the original KB, and word features) and the Tensorflow
backend, with minibatches of size 100 and an Adagrad optimizer with a
learning rate of 0.1, running for 20 epochs, and no regularization.
We compare accuracy results with two prior neural-network based
methods which have been applied to this task.  As shown in
Table~\ref{tab:wikimovies}, TensorLog performs better than the prior
state-of-the-art on this task, and is quite efficient.

\section{Concluding Remarks}

In this paper, we described a scheme to integrate probabilistic
logical reasoning with the powerful infrastructure that has been
developed for deep learning.  The end goal is to enable deep learners
to incorporate first-order probabilistic KBs, and conversely, to
enable probabilistic reasoning over the outputs of deep learners.
TensorLog, the system we describe here, makes this possible to do at
reasonable scale using conventional neural-network platforms.

This paper contains several interrelated technical
contributions. First, we identified a family of probabilistic
deductive databases (PrDDBs) called {polytree-limited stochastic
  deductive knowledge graphs (ptree-SDKGs)} which are tractable, but
still reasonably expressive.  This language is a variant of SLPs, and
it is maximally expressive, in that one cannot drop the polytree
restriction, or switch to a possible-worlds semantics, without making
inference intractible.  We argue above that logics which are not
tractable (i.e., are \#P or worse in complexity) are unlikely to be
practically incorporated into neural networks.

Second, we presented an algorithm for performing inference for
ptree-SDKGs, based on belief propagation.  Computationally, the
algorithm is quite efficient. Assuming the matrices $\M_p$ exist, the
additional memory needed for the factor-graph $G_r$ is linear in the
size of the clause $r$, and hence the compilation is linear in the
theory size and recursion depth.  To our knowledge use of BP for
first-order inference in this setting is novel.

Finally, we present an implementation of this logic, called TensorLog.
The implementation makes it possible to both call TensorLog inference
within neural models, or conversely, to call neural models within
TensorLog.

The current implementation of TensorLog includes a number of
restrictions.  Two backends are implemented, one for Tensorflow and
one for Theano, but the Tensorflow backend has been more extensively
tested and evaluated.  We are also exploring compilation to
PyTorch\footnote{pytorch.org}, which supports dynamic networks.  We
also plan to implement support for more stable optimization (e.g.,
gradient clipping), and better support for debugging.  

As noted above, TensorLog also makes it possible to replace components
of the logic program (e.g., the \cd{classification} or \cd{matches}
predicate) with submodels learned in the deep-learning infrastructure.
Alternatively, one can export a \cd{answer} predicate defined by the
logic to a deep learner, as a function which maps a question to
possible answers and their confidences; this might be useful in
building a still more complex model non-logical model (e.g., a dialog
agent which makes use of question-answering as a subroutine.)  In
future work we hope to explore these capabilities.

We also note that although the experiments in this paper assume that
theories are given, the problem of learning programs in TensorLog is
also of great interest. Some early results from the authors on this
problem are discussed elsewhere \cite{yang2017differentiable}.

\acks{Thanks to William Wang for providing some of the datasets used
  here; and to William Wang and many other colleagues contributed with
  technical discussions and advice.  The author is greatful to Google
  for financial support, and also to NSF for their support of his work
  via grants CCF-1414030 and IIS-1250956.}

\newpage

\appendix

\section{Proofs}

\setcounter{thm}{0}
\begin{thm}
\statementone
\end{thm}

We will reduce counting proofs for 2PSAT to computing probabilities
for SDKGs.  2PSAT is a \#P-hard task where the goal is to count the
number of satisfying assignments to a CNF formula with only two
literals per clause, all of which are positive
\cite{suciu2011probabilistic}. Hence a 2PSAT formula is of the form
\[ (x_{a_1} \vee x_{b_1}) \wedge \ldots \wedge (x_{a_n} \vee \ell_{b_n})
\]
where the variables are all binary variables $x_i$ from $X =
\{x_1,\ldots,x_n\}$, and each $a_i$ and $b_i$ is a index between 1 and
$n$.  The subformula $(x_{a_i} \vee x_{b_i})$ is called the $i$-th
clause below.

%Consider the four possible assignments 00, 01, 10, and 11 to the two
%variables $x_{a_i},x_{b_i}$.  The $i$-th clause will be made true for
%any of these except for 00.  We encode this information in the
%following database facts, all with weight 1.
%
%\begin{tabular}{lll}
%sat\_assign(01) & bit1(01,0)      & bit2(01,1) \\
%sat\_assign(10) & bit1(10,1)      & bit2(10,0) \\
%sat\_assign(11) & bit1(11,1)      & bit2(11,1) \\
%\end{tabular}

The database contains the two facts \cd{assign\_yes(yes)} and
\cd{assign\_no(no)} with weight 1, and two facts \cd{binary(0)} and
\cd{binary(1)} with weights 0.5.  It also contains a definition of the
predicate \cd{either\_of}, containing the following three weight 1
facts: \cd{either\_of(0,1)}, \cd{either\_of(1,0)}, and
\cd{either\_of(1,1)}.  There are a total of 7 facts in the database.

%\begin{tabbing}123456\=\kill
%\cd{sat(Y)} :- classification(Q, \\
%\>1) \cd{assign\_(Y), binary(X$_1$), \ldots, binary(X$_n$),} \\
%\>   \cd{sat\_assign(A$_1$), bit1(A$_1$,X$_{a_1}$), bit2(A$_1$,X$_{b_1}$),} \\
%\>   \ldots,\\
%\>2) \cd{sat\_assign(A$_i$), bit1(A$_i$,X$_{a_i}$), bit2(A$_i$,X$_{b_i}$),} \\
%\>   \ldots,\\
%\>   \cd{sat\_assign(A$_n$), bit1(A$_n$,X$_{a_n}$), bit2(A$_n$,X$_{b_n}$),} \\
%\cd{sat(Y) :-} \\
%\>   \cd{assign\_no(Y)}.
%\end{tabbing}

\begin{tabbing}123456\=\kill
\cd{sat(Y)} :- \\
\> \cd{assign\_yes(Y), binary(X$_1$), \ldots, binary(X$_n$),} \\
\>   \cd{either\_of(X$_{a_1}$,X$_{b_1}$)}, \\
\>   \ldots,\\
\>   \cd{either\_of(X$_{a_n}$,X$_{b_n}$)}, \\
\cd{sat(Y) :-} \\
\>   \cd{assign\_no(Y)}.
\end{tabbing}

In the first line of the first rule, an assignment to the $x_i$'s is
selected, with uniform probability.  It is easy to see that the
literal \cd{either\_of(X$_{a_i}$,X$_{b_i}$)} will succeed iff the
$i$-th clause is made true by this assignment.  Hence the first rule
of the theory will succeed exactly $k$ times, where $k$ is the number
of satisfying assignments for the formula.  The second clause succeeds
once, so
\[ p = \Pr(\cd{sat(yes)}|\cd{sat(Y)}) = \frac{k}{k+1}
\]  
If $p$ could be computed efficiently, one could solve the equation
above for $k$ and use the result to determine the number of satisfying
assignments to the 2PSAT formula.

\setcounter{thm}{2}
\begin{thm}
\statementthree
\end{thm}

We again reduce counting assignments for 2NSAT to computation of $p =
\Pr(\cd{sat(yes)})$.  In this case the DB contains $n$ facts of the
form \cd{x$_1$(1)}, \cd{x$_2$(1)}, \ldots, \cd{x$_n$(1)}, all with
weights 0.5, and the additional fact \cd{assign\_yes(yes)}.

We can now encode the 2PSAT formula with the following theory.  For
each clause $i$ let $j1$ and $j2$ be the indices of the two literals
in that clause. We construct two theory rules for each clause $i$:
\begin{tabbing}123456\=\kill
  \cd{sat$_i$(Y) :- x$_{j1}$(Y).}\\
  \cd{sat$_i$(Y) :- x$_{j2}$(Y).} \\
\end{tabbing}
Finally we add a binary tree of $O(\log(n))$ rules, each of which test
success of two other subpredicates, and the last of which tests
succeeds only of all the \cd{clause}$_i$ predicates succeed.  For
instance, for $n=8$, we would define \cd{sat(Y)} as
\begin{tabbing}123456\=\kill
\cd{sat$_{1:2}$(Y) :- clause$_1$(Y),clause$_2$(Y).}\\
\cd{sat$_{2:3}$(Y) :- clause$_2$(Y),clause$_3$(Y).}\\
\cd{sat$_{1:4}$(Y) :- sat$_{1:2}$(Y),sat$_{2:3}$(Y).} \\
\cd{sat$_{5:6}$(Y) :- clause$_5$(Y),clause$_6$(Y).}\\
\cd{sat$_{7:8}$(Y) :- clause$_7$(Y),clause$_8$(Y).}\\
\cd{sat$_{5:8}$(Y) :- sat$_{5:6}$(Y),sat$_{7:8}$(Y).} \\
\cd{sat(Y) :- sat$_{1:4}$(Y),sat$_{5:8}$(Y),assign\_yes(Y).} \\
\end{tabbing}
Note that for each variable $x_j$, an $I$ drawn from the database
distribution may contain either \cd{x$_j$(1)} or not, so there are
$2^n$ possible interpretations. Each of these corresponds to a
boolean assignment, where $x_j=1$ in the assignment exactly when
\cd{x$_j$(1)} is in the corresponding interpretation.  Clearly the
\cd{clause}$_i$ predicate succeeds exactly when the $i$-th clause is
satisfied, and hence $p = \Pr(\cd{sat(yes)} | \DB,|T)$ is thus exactly
$\frac{k}{2^n}$, where $k$ is the number of satisfying assignments.

This theory is quite simple: it contains no binary predicates, and (if
one tests success of all clause predicates in a tree) the rules are
all very short.  It is thus difficult to identify syntactic
restrictions which might make proof-counting tractible for the
possible-worlds scenario.

\bibliography{all}
\bibliographystyle{theapa}

\end{document}